\documentclass[10pt,twocolumn,letterpaper]{article}

\usepackage{wacv}
\usepackage{xcolor}
\usepackage{times}
\usepackage{epsfig}
\usepackage{graphicx}
\usepackage{amsmath}
\usepackage{mathtools}
\usepackage{amssymb}
\usepackage{caption}
\usepackage{subcaption}
\usepackage[normalem]{ulem}

\def\wacvPaperID{903} %

\wacvfinalcopy %

\ifwacvfinal
\fi

\ifwacvfinal
\usepackage[breaklinks=true,bookmarks=false]{hyperref}
\else
\usepackage[pagebackref=true,breaklinks=true,colorlinks,bookmarks=false]{hyperref}
\fi

\newcommand{\ourstudent}{entropic student\xspace}
\newcommand{\uourstudent}{Entropic Student\xspace}
\newcommand{\repourl}{\url{https://github.com/yoshitomo-matsubara/supervised-compression}}

\DeclarePairedDelimiter\round{\lfloor}{\rceil}

\def\z{\mathbf{z}}
\def\h{\mathbf{h}}
\def\hhat{\mathbf{\hat{h}}}
\def\y{\mathbf{y}}
\def\x{\mathbf{x}}

\wacvfinalcopy

\begin{document}

\title{Supervised Compression for Resource-Constrained Edge Computing Systems}

\author{Yoshitomo Matsubara ~~~~~~ Ruihan Yang ~~~~~~ Marco Levorato ~~~~~~ Stephan Mandt\\
Department of Computer Science, University of California, Irvine\\
{\tt\small \{yoshitom, ruihan.yang, levorato, mandt\}@uci.edu}
}

\maketitle

\begin{abstract}
There has been much interest in deploying deep learning algorithms on low-powered devices, including smartphones, drones, and medical sensors. However, full-scale deep neural networks are often too resource-intensive in terms of energy and storage. As a result, the bulk part of the machine learning operation is therefore often carried out on an edge server, where the data is compressed and transmitted. However, compressing data (such as images) leads to transmitting information irrelevant to the supervised task. Another popular approach is to split the deep network between the device and the server while compressing intermediate features. To date, however, such split computing strategies have barely outperformed the aforementioned naive data compression baselines due to their inefficient approaches to feature compression. This paper adopts ideas from knowledge distillation and neural image compression to compress intermediate feature representations more efficiently. Our supervised compression approach uses a teacher model and a student model with a stochastic bottleneck and learnable prior for entropy coding (\uourstudent). We compare our approach to various neural image and feature compression baselines in three vision tasks and found that it achieves better supervised rate-distortion performance while maintaining smaller end-to-end latency. We furthermore show that the learned feature representations can be tuned to serve multiple downstream tasks. 
\end{abstract}

\section{Introduction}
\label{sec:intro}
With the abundance of smartphones, autonomous drones, and other intelligent devices, advanced computing systems for machine learning applications have become evermore important~\cite{shi2016edge,chen2019deep}.
Machine learning models are frequently deployed on low-powered devices for reasons of computational efficiency or data privacy~\cite{sandler2018mobilenetv2,howard2019searching}.
However, deploying conventional computer vision or NLP models on such hardware raises a computational challenge, as powerful deep neural networks are often too energy-consuming to be deployed on such weak mobile devices~\cite{eshratifar2019jointdnn}. 

\begin{figure}[t]
    \centering
    \includegraphics[width=1.0\linewidth]{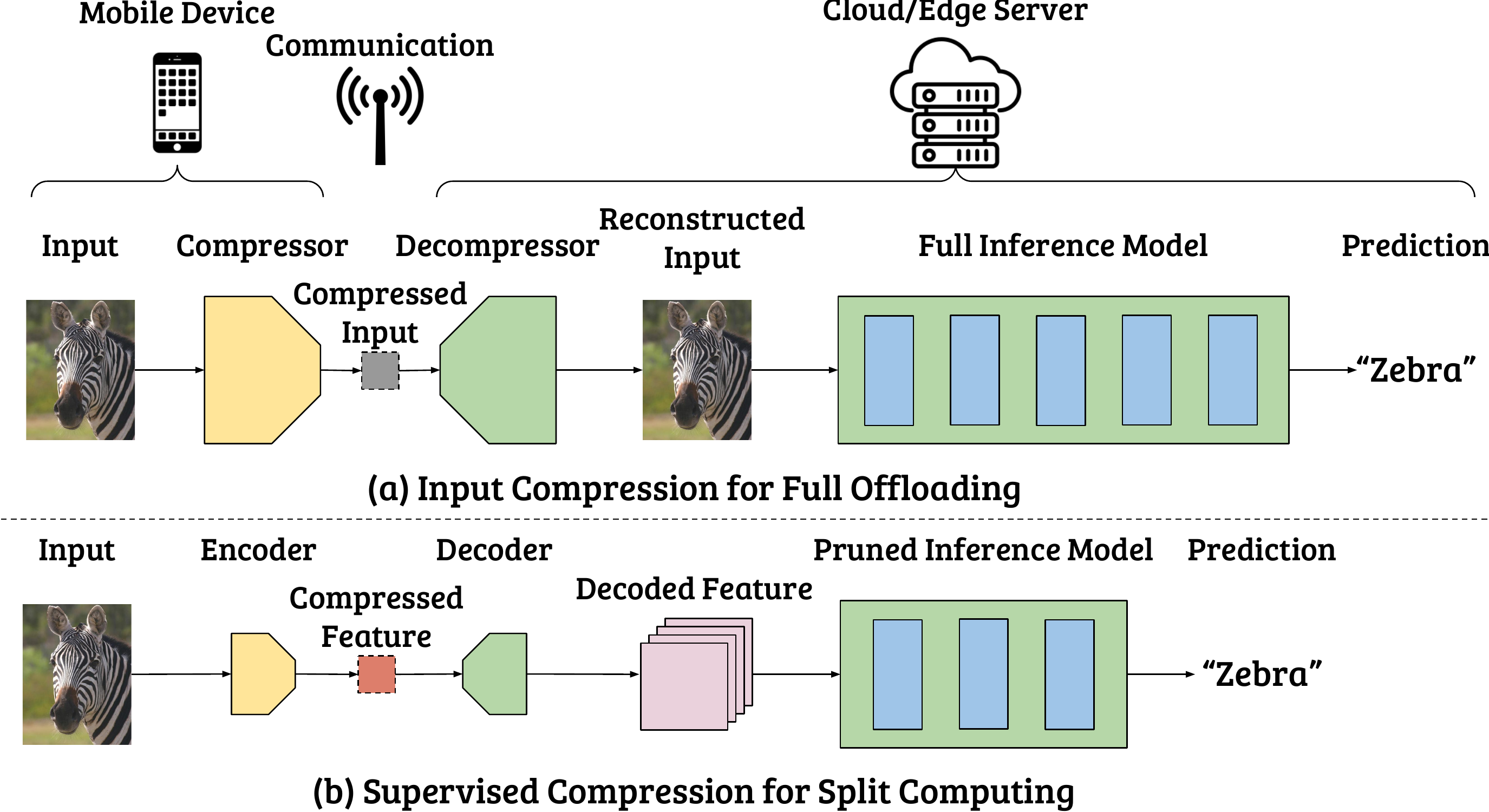}
    \vspace{-2em}
    \caption{Image classification with input compression (\textbf{top}) vs. our proposed supervised compression for split computing (\textbf{bottom}). While the former approach fully reconstructs the image, our approach learns an intermediate compressible representation suitable for the supervised task.}
    \label{fig:input_comp_vs_feat_comp}
\end{figure}

An alternative to carrying out the deep learning model's operation on the low-powered device is to send compressed data to an edge server that takes care of the heavy computation instead.
However, recent neural or classical compression algorithms are either resource-intensive~\cite{singh2020end,dubois2021lossy} and/or optimized for perceptual quality~\cite{balle2017end,minnen2018joint}, and therefore much of the information transmitted is redundant for the machine learning task~\cite{choi2020task} (see Fig.~\ref{fig:input_comp_vs_feat_comp}). A better solution is to therefore split the neural network~\cite{kang2017neurosurgeon} into the two sequences so that some elementary feature transformations are applied by the first sequence of the model on the weak mobile (local) device.
Then, intermediate, informative features are transmitted through a wireless communication channel to the edge server that processes the bulk part of the computation (the second sequence of the model)~\cite{eshratifar2019bottlenet,matsubara2019distilled}. 

Traditional split computing approaches transmit intermediate features by either reducing channels in convolution layers~\cite{kang2017neurosurgeon} or truncating them to a lower arithmetic precision~\cite{matsubara2020head,shao2020bottlenet++,matsubara2020neural}. Since the models were not ``informed'' about such truncation steps during training oftentimes leads to substantial performance degradation. This raises the question of whether \emph{learnable} end-to-end data compression pipelines can be designed to both truncate \emph{and} entropy-code the involved early-stage features. 

In this paper, we propose such a neural feature compression approach by drawing on variational inference-based data compression~\cite{balle2018variational,singh2020end}. Our architecture resembles the variational information bottleneck objective~\cite{alemi2016deep} and relies on an encoder, a ``prior'' on the bottleneck state, and a decoder that leads to a supervised loss (see Fig.~\ref{fig:input_comp_vs_feat_comp}).
At inference time, we discretize the encoder's output and use our learned prior as a model for entropy coding intermediate features.
The decoder reconstructs the feature vector losslessly from the binary bitstring and carries out the subsequent supervised machine learning task. Crucially, we combine this feature compression approach with knowledge distillation, where a teacher model provides the training data as well as parts of the trained architecture.\footnote{Code and models are available at \repourl} 
    
In more detail, our main contributions are as follows:
\vspace{-0.5em}
\begin{itemize}
    \item We propose a new training objective for feature compression in split computing that allows us to use a learned entropy model for bottleneck quantization in conjunction with knowledge distillation. 
    \item Our approach significantly outperforms seven strong baselines from the split computing and (neural) image compression literature in terms of rate-distortion performance (with distortion measuring a supervised error) and in terms of end-to-end latency. 
    \item Moreover, we show that a single encoder network can serve multiple supervised tasks, including classification, object detection, and semantic segmentation.  
\end{itemize}

\section{Related Work}
\label{sec:related}

\paragraph{Neural Image Compression.} Neural image compression methods apply neural networks for nonlinear dimensionality reduction and subsequent entropy coding. Early works~\cite{toderici2017full,johnston2018improved} leveraged LSTMs to model spatial correlations of the pixels within an image. The first proposed autoencoder architecture for image compression~\cite{theis2017lossy} used the straight-through estimator~\cite{bengio2013estimating} for learning a discrete latent representation. The connection of image compression to \emph{probabilistic} generative models was drawn by variational autoencoders (VAEs)~\cite{kingma2013auto, balle2017end}. In the subsequent work \cite{balle2018variational},  two-level VAE architectures involving a scale hyper-prior are proposed to encode images, which can be further improved by autoregressive structures~\cite{minnen2018joint,minnen2020channel} or by optimization at encoding time~\cite{yang2020improving}. Recent work also shows the potential progressive compression of the VAE structure by extending the quantization grid~\cite{lu2021progressive}. Other works~\cite{yang2020variational,flamich2019compression} demonstrate competitive image compression performance without a pre-defined quantization grid.

Recently, Dubois \emph{et al.}~\cite{dubois2021lossy} propose a self-supervised compression architecture for generic image classification.
However, their encoder involves 87.8 million parameters (627 times larger than our encoder in Table~\ref{table:model_size}) due to its Vision Transformer (ViT~\cite{dosovitskiy2020image})-based encoder used in CLIP model~\cite{radford2021learning} for ImageNet dataset.
Thus, it does not satisfy resource-constrained edge computing systems.

\paragraph{Split Computing.} 
Given that mobile (or local) devices often have limited resources such as computing power and battery, we usually transfer sensor data captured by the mobile device and offload heavy computing tasks to an edge (or cloud) server with more computing resources.
Unlike local computing, which executes the entire model on the mobile device, edge computing (or full offloading) where the computation is on the edge server requires quality wireless communication between the mobile device and edge server.
Otherwise, the total inference cost, such as the end-to-end latency, would be higher than local computing due to the communication delay, which would be critical for real-time applications.
As an intermediate option between local computing and edge computing, split computing~\cite{kang2017neurosurgeon} has been attracting attention from research communities since edge computing is not always the best option.
Specifically, the communication cost would be a severe problem for resource-limited edge computing systems~\cite{eshratifar2019bottlenet,matsubara2019distilled}.

In split computing, a neural model will be split into the first and second sequences, and the first sequence of the model is executed on the mobile device.
Having received the output of the first section via wireless communication, the second sequence of the model completes the inference on the edge server.
A key concept is to reduce computational load on the mobile device while saving communication cost (data size) as processing delay on the edge server is often smaller compared to local processing and communication delays~\cite{matsubara2020neural}.
For reducing communication cost, recent studies on split computing~\cite{eshratifar2019bottlenet,matsubara2019distilled,shao2020bottlenet++,matsubara2020neural} introduce \emph{bottleneck}, whose data size is smaller than input sample, to vision models.
In recent studies, a combination of 1) fewer channels (channel reduction) in convolution layers and 2) quantization at bottleneck point is key to design such bottlenecks.
While such studies show the effectiveness of their proposed approach for image classification and object detection tasks, the accuracy of bottleneck-injected models sharply drops when further compressing the bottleneck size as we will describe in Section~\ref{sec:experiments}.

\paragraph{Knowledge Distillation.} 
It is widely known that deep neural models are often overparameterized~\cite{ba2014do,urban2016do}, and knowledge distillation~\cite{hinton2014distilling} is one of the well-known techniques for model compression~\cite{bucilua2006model}.
In the paradigm, a large pretrained model plays a role of \emph{teacher} for a model to be trained (called \emph{student}), and the student model learns from both \emph{hard}-targets (\emph{e.g.}, one-hot vectors) and \emph{soft}-targets (outputs of the teacher for the given input) during training.
Interestingly, some uncertainty from the pretrained teacher model as \emph{soft}-target is informative to student models. The models trained with teachers often achieve better prediction performance than those trained without teachers~\cite{ba2014do}.
As will be discussed later in this paper, learning compressed features for a target task such as image classification~\cite{singh2020end} would be challenging, specifically in the case that we introduce such bottlenecks to early layers in the model.
We leverage a pretrained model as the teacher, and those with introduced bottlenecks as students to be trained to improve rate-distortion performance.

\section{Method}
\label{sec:method}
After providing an overview of the setup (Section~\ref{sec:overview}) we describe our distillation and feature compression approach (Section~\ref{sec:distillation}) and our procedure to fine-tune the model to other supervised downstream tasks (Section~\ref{sec:finetune}).

\subsection{Overview}
\label{sec:overview}
Our goal is to learn a lightweight, communication-efficient feature extractor for supervised downstream applications. We thereby transmit intermediate feature activations between two distinct portions of a neural network. The first part is deployed on a low-power mobile device, and the second part on a compute-capable edge server. Intermediate feature representations are compressed and transmitted between the mobile device and the edge server using discretization and subsequent lossless entropy coding. 

In order to learn good compressible feature representations, we combine two ideas: \emph{knowledge distillation} and neural data compression via \emph{learned entropy coding}. First, we train a large teacher network on a data set of interest to teach a smaller student model. We assume that the features that the teacher model learns are helpful for other downstream tasks. Then, we train a lightweight student model to match the teacher model's intermediate features (Section~\ref{sec:distillation}) with minimal performance loss. Finally, we fine-tune the student model to different downstream tasks (Section~\ref{sec:finetune}). 
Note that the training process is done offline.

The teacher network realizes a deterministic mapping $\x \mapsto \h \mapsto \y$, where $\x$ are the input data, $\y$ are the targets, and $\h$ are some intermediate feature representations of the teacher network. We assume that the teacher model is too large to be executed on the mobile device. The main idea is to replace the teacher model's mapping $\x \mapsto \h$ with a student model  (\emph{i.e.}, the new targets become the teacher model's intermediate feature activations). To facilitate the data transmission from the mobile device to the edge server, the student model, \emph{\uourstudent}, embeds a bottleneck representation $\z$ that allows compression (see details below), and we transmit data as $\x \mapsto \z \mapsto \h \mapsto \y$. 
We show that the student model can be fine-tuned to different tasks while the mobile device's encoder part remains unchanged. 

The whole pipeline is visualized in the bottom panel of Fig.~\ref{fig:input_comp_vs_feat_comp}.
The latent representation $\z$ has a ``prior'' $p(\z)$, \emph{i.e.}, a density model over the latent space $\z$ that both sender and receiver can use for entropy coding after discretizing $\z$. In the following, we derive the details of the approach.  

\subsection{Knowledge Distillation}
\label{sec:distillation}

\begin{figure}[t]
    \centering
    \includegraphics[width=0.95\linewidth]{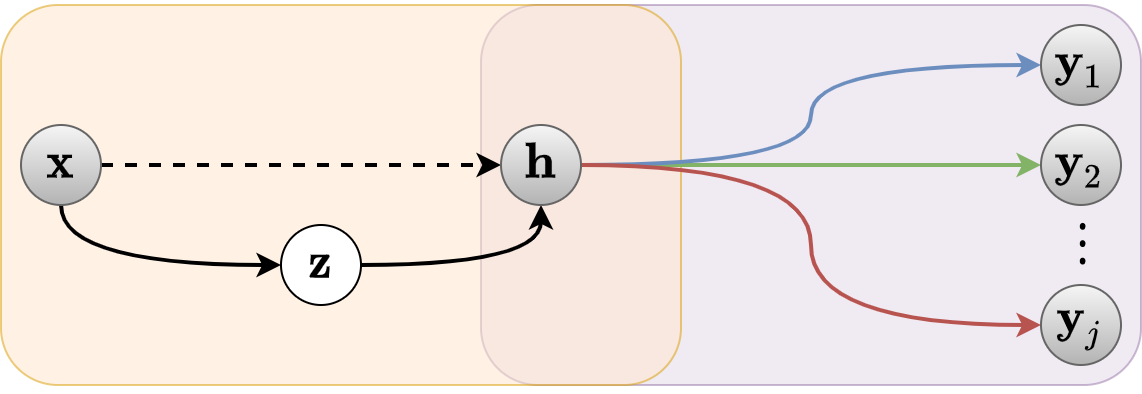}
    \vspace{-0.5em}
    \caption{Proposed graphical model. Black arrows indicate the compression and decompression process in our student model. The dashed arrow shows the teacher's original deterministic mapping. Colored arrows show the discriminative tail portions shared between student and teacher.}
    \label{fig:prob_graph}
\end{figure}

We first focus on the details of the distillation process.
The \ourstudent model learns the mapping $\x \mapsto \h$ (Fig.~\ref{fig:prob_graph}, left part) by drawing samples from the teacher model.
The second part of the pipeline $\h \mapsto \y$ (Fig.~\ref{fig:prob_graph}, right part) will be adapted from the teacher model and will be fine-tuned to different tasks (see Section~\ref{sec:finetune}).

Similar to neural image compression~\cite{balle2017end,balle2018variational}, we draw on latent variable models whose latent states allow us to quantize and entropy-code data under a prior probability model. In contrast to neural image compression, our approach is supervised. As such, it mathematically resembles the deep Variational Information Bottleneck~\cite{alemi2016deep} (which was designed for adversarial robustness rather than compression). 

\paragraph{Distillation Objective.}
\begin{figure*}[t]
    \centering
    \includegraphics[width=0.9\linewidth]{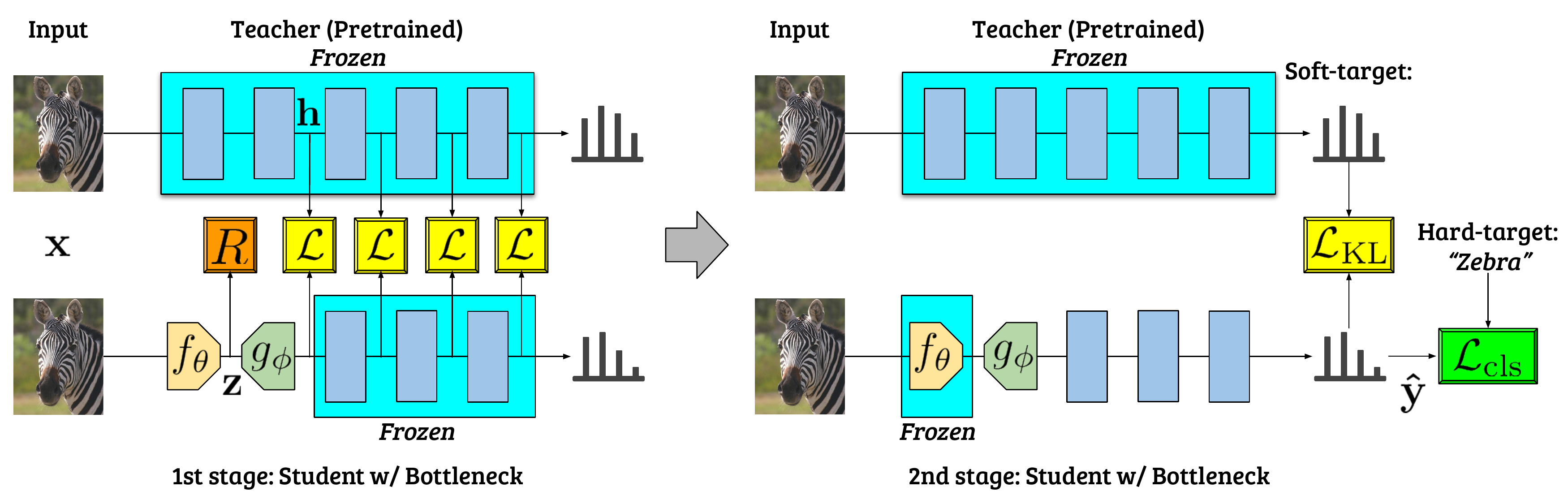}
    \vspace{-1em}
    \caption{Our two-stage training approach. \textbf{Left}: training the student model (\textbf{bottom}) with targets $\h$ and tail architecture obtained from teacher (\textbf{top}) (Section~\ref{sec:distillation}). \textbf{Right}: fine-tuning the decoder and tail portion with fixed encoder (Section~\ref{sec:finetune}).}
    \label{fig:proposed_train}
\end{figure*}

We assume a stochastic encoder $q(\z|\x)$, a decoder $p(\h|\z)$, and a density model (``prior'') $p(\z)$ in the latent space. Specific choices are detailed below. Similar to~\cite{alemi2016deep}, we \emph{maximize} mutual information between $\z$ and $\h$ (making the compressed bottleneck state $\z$ as informative as possible about the supervised target $\h$) while \emph{minimizing} the mutual information  between the input $\x$ and $\z$ (thus ``compressing away'' all the irrelevant information that does not immediately serve the supervised goal). 

The objective for a given training pair $(\x,\h)$ provided by the teacher model is 
\begin{align}
    \mathcal{L}(\x,\h) = -  \underbrace{\mathbb{E}_{q_\theta(\z|\x)}[\log p_\phi(\h|\z)}_{\text{distortion}} + \beta \underbrace{\log p_\phi(\z)}_{\text{rate}}]. %
    \label{eq:supervised-RD}
\end{align}
Before discussing the rate and distortion terms, we specify and simplify this loss function further. Above, the decoder $p(\h|\z) = {\cal N}(\h; g_\phi(\z), I)$ is chosen as a conditional Gaussian centered around a deterministic prediction $g_\phi(\z)$. 
Following the neural image compression literature~\cite{balle2017end,balle2018variational}, the \emph{encoder} is chosen to be a unit-width box function $q_\theta(\z|\x) = \mathcal{U}(f_\theta(\x)-\frac{1}{2}, f_\theta(\x)+\frac{1}{2})$ centered around a neural network prediction $f_\theta(\x)$. Using the reparameterization trick~\cite{kingma2013auto}, Eq.~\ref{eq:supervised-RD} can be optimized via stochastic gradient descent~\footnote{For better convergence, we follow~\cite{matsubara2020neural} and leverage intermediate representations from frozen layers besides $\h$ as illustrated in Fig.~\ref{fig:proposed_train} (left).}
\begin{multline}
    \mathcal{L}(\x,\h) = \textstyle{\frac{1}{2}}  \underbrace{|| \h - g_{\phi}\left(f_{\theta}\left(\mathbf{x}\right) + \epsilon \right) ||_{2}^{2}}_\text{distortion} \\ - \beta \underbrace{\log p_\phi(f_\theta(\x) + \epsilon)}_{\text{rate}}, \quad \epsilon \sim \text{Unif}(\textstyle{-\frac{1}{2},\frac{1}{2}}).
    \label{eq:proposed_1st}
\end{multline}
Once the model is trained, we discretize the latent state $\z = \round{f_\theta(\x)}$ (where $\round{\cdot}$ denotes the rounding operation) to allow for entropy coding under $p_\phi(\z)$. 
By injecting noise from a box-shaped distribution of width one, we simulate the rounding operation during training. The entropy model or prior $p_\phi(\z)$ is adopted from the neural image compression literature~\cite{balle2018variational}; it is a learnable prior with tuning parameters $\phi$. The prior factorizes over all dimensions of $\z$, allowing for efficient and parallel entropy coding.

\paragraph{Supervised Rate-Distortion Tradeoff.}
Similar to unsupervised data compression, our approach results in a rate-distortion tradeoff: the more aggressively we compress the latent representation $\z$, the more the predictive performance will deteriorate. In contrast, the more bits we are willing to invest for compressing $\z$, the more predictive strength our model will maintain. The goal will be to perform well on the unavailable tradeoff between rate and distortion. 

The first term in Eq.~\ref{eq:supervised-RD} measures the supervised distortion, as it expresses the average prediction error under the coding procedure of first mapping $\x$ to $\z$ and then $\z$ to $\h$. In contrast, the second term measures the coding costs as the cross-entropy between the empirical distribution of $\z_i$ and the prior distribution $p_\phi(\z)$ according to information theory~\cite{cover1999elements}. The tradeoff is determined by the Lagrange multiplier $\beta$.  

As a particular instantiation of an information bottleneck framework~\cite{alemi2016deep}, Singh \emph{et al.}~\cite{singh2020end} proposed a similar loss function as Eq.~\ref{eq:supervised-RD} to train a classifier with a bottleneck at its penultimate layer without knowledge distillation.
In Section~\ref{sec:experiments}, we compare against a version of this approach that is compatible with our architecture and find that the knowledge distillation aspect is crucial to improve performance.

\subsection{Fine-tuning for Target Tasks}
\label{sec:finetune}
Equation~\ref{eq:supervised-RD} shows the base approach, describing the knowledge distillation pipeline with a single $\h$ and involving a single target $\y$. In practice, our goal is to learn a compressed representation $\z$ that does not only serve a single supervised target $\y$, but multiple ones $\y_1, \cdots, \y_j$. In particular, for a deployed system with a learned compression module, we would like to be able to fine-tune the part of the network living on the edge server to multiple tasks without having to retrain the compression model. As follows, we show that such multi-task learning is possible. 

A learned student model from knowledge distillation can be depicted as a two-step deterministic mapping $\z=\round{f_\theta(\x)}$ and $\hhat = g_\phi(\z)$, where $\hhat$ ($\approx \h$) is now a decompressed intermediate hidden feature in our final student model (see Fig.~\ref{fig:prob_graph}). Assuming that $p_{\psi_j}(\y_j|\hhat)$ denotes the student model's output probability distribution with parameters $\psi_j$, the fine-tuning step amounts to optimizing
\begin{equation}
    \psi_j^* = \arg\min_{\psi_j} - \mathbb{E}_{(\x,\y)\sim\mathcal{D}} [p_{\psi_j}(\y_j|g_\phi(\round{f_\theta(\x)}))].
\end{equation}
The pair $(\y_j, \psi_j)$ refers to the target label and the parameters of each downstream task. The formula illustrates the Maximum Likelihood Estimation (MLE) method to optimize the parameter $\psi_j$ for task $j$.
Note that we optimize the discriminative model after the compression model is frozen, so $\theta$ is fixed in this training stage, and $\phi$ can either be fixed or trainable. We elucidate the hybrid model in Fig.~\ref{fig:prob_graph}.

For fine-tuning the student model with the frozen encoder, we leverage a teacher model again.
For image classification, we apply a standard knowledge distillation technique~\cite{hinton2014distilling} to achieve better model accuracy by distilling the knowledge in the teacher model into our student model.
Specifically, we fine-tune the student model by minimizing a weighted sum of two losses: 1) cross-entropy loss between the student model's class probability distribution and one-hot vector (\emph{hard}-target), and 2) Kullback-Leibler divergence between \emph{softened} class probability distributions from both the student and teacher models.

Similarly, having frozen the encoder, we can fine-tune different models for different downstream tasks reusing the trained \ourstudent model (classifier) as their backbone, which will be demonstrated in Section~\ref{subsec:additional_tasks}.

\section{Experiments}
\label{sec:experiments}
Using torchdistill~\cite{matsubara2021torchdistill}, we designed different experiments and studied various models based on both principles of split-computing (partial offloading) and edge computing (full offloading).
We used ResNet-50~\cite{he2016deep} as a base model, which, besides image classification, is also widely used as a backbone for different vision tasks such as object detection~\cite{he2017mask,tsung2017focal} and semantic segmentation~\cite{chen2017rethinking}.
In all experiments, we empirically show that our approach leads to better supervised rate-distortion performance. 

\subsection{Baselines}
In this study, we use seven baseline methods categorized into either input compression or feature compression.

\vspace{-1em}
\paragraph{Input compression (IC).}
A conventional implementation of the edge computing paradigm is to transmit the compressed image directly to the edge server, where all the tasks are then executed. We consider five baselines referring to this ``input compression'' scenario: JPEG, WebP~\cite{webp}, BPG~\cite{bpg}, and two neural image compression methods (factorized prior and mean-scale hyperprior)~\cite{balle2018variational,minnen2018joint} based on CompressAI~\cite{begaint2020compressai}.
The latter approach is currently considered state of the art in image compression models (without autoregressive structure)~\cite{minnen2018joint,minnen2020channel,yang2020improving}.
We evaluate each model's performance in terms of the rate-distortion curve by setting different quality values for JPEG, WebP, and BPG and Lagrange multiplier $\beta$ for neural image compression.

\vspace{-1em}
\paragraph{Feature compression (FC).}
Split computing baselines~\cite{matsubara2020head,shao2020bottlenet++} correspond to reducing the bottleneck data size with channel reduction and bottleneck quantization referred to as CR+BQ (quantizes 32-bit floating-point to 8-bit integer)~\cite{jacob2018quantization}. Matsubara \emph{et al.}~\cite{matsubara2020neural,matsubara2020head} report that bottleneck quantization did not lead to significant accuracy loss. To control the rate-distortion tradeoff, we design bottlenecks with a different number of output channels in a convolution layer to control the bottleneck data size, train the bottleneck-injected models and quantize the bottleneck after the training session.

Our final baseline in this paper is an end-to-end approach towards learning compressible features for a single task similar to Singh \emph{et al.}~\cite{singh2020end} (for brevity, we will cite their reference). Their originally proposed approach focuses only on classification and introduces the compressible bottleneck to the penultimate layer. In the considered setting, such design leads to an overwhelming workload for the mobile/local device: for example, in terms of model parameters, about 92\% of the ResNet-50~\cite{he2016deep} parameters would be deployed on the weaker, mobile device. To make this approach compatible with our setting, we apply their approach to our architecture; that is, we directly train our \ourstudent model without a teacher model. We find that compared to~\cite{singh2020end}, having a stochastic bottleneck at an earlier layer (due to limited capacity of mobile devices) leads to a model that is much harder to optimize (see Section~\ref{subsec:imagenet}).

\subsection{Implementation of Our \uourstudent}
\label{subsec:student_model}
Vision models in recent years reuse pretrained image classification models as their backbones \emph{e.g.}, ResNet-50~\cite{he2016deep} as a backbone of RetinaNet~\cite{tsung2017focal} and Faster R-CNN~\cite{ren2015faster} for object detection tasks.
These models often use intermediate hidden features extracted from multiple layers in the backbone as the input to subsequent task-specific modules such as feature pyramid network (FPN)~\cite{lin2017feature}.
Thus, using an architecture with a bottleneck introduced at late layers~\cite{singh2020end} for tasks other than image classification may require transferring and compressing multiple hidden features to an edge server, which will result in high communication costs.

To improve the efficiency of split computing compared to that of edge computing, we introduce the bottleneck as early in the model as possible to reduce the computational workload at the mobile device.
We replace the first layers of our pretrained teacher model with the new modules for encoding and decoding transforms as illustrated in Fig.~\ref{fig:proposed_train}.
The student model, \emph{\ourstudent}, consists of the new modules and the remaining layers copied from its teacher model for initialization.
Similar to neural image compression models~\cite{balle2018variational,minnen2018joint}, we use convolution layers and simplified generalized divisive normalization (GDN) \cite{balle2015density} layers to design an encoder $f_\theta$, and design a decoder $g_\phi$ with convolution and inverse version of simplified GDN (IGDN) layers.
Importantly, the designed encoder should be lightweight, \emph{e.g.}, with fewer model parameters as it will be deployed and executed on a low-powered mobile device.
We will discuss the deployment cost in Section~\ref{subsec:model_size}.

Different from bottleneck designs in the prior studies on split computing~\cite{matsubara2020neural,matsubara2020head}, we control the trade-off between bottleneck data size and model accuracy with the $\beta$ value in the rate-distortion loss function (See Eq.~\ref{eq:proposed_1st}).

\subsection{Image Classification}
\label{subsec:imagenet}
We first discuss the rate-distortion performance of our and baseline models using a large-scale image classification dataset.
Specifically, we use ImageNet (ILSVRC 2012)~\cite{russakovsky2015imagenet}, that consists of 1.28 million training and 50,000 validation samples.
As is standard, we train the models on the training split and report the top-1 accuracy on the validation split.
Using ResNet-50~\cite{he2016deep} pre-trained on ImageNet as a teacher model, we replace all the layers before its second residual block with our encoder and decoder to compose our \emph{\ourstudent} model.
The introduced encoder-decoder modules are trained to approximate $\h$ in Eq.~\ref{eq:proposed_1st}, which is the output of the corresponding residual block in the teacher model (original ResNet-50) in the first stage, and then we fine-tune the student model as described in Section~\ref{sec:method}.
We provide more details of training configurations (\emph{e.g.}, hyperparameters) in the supplementary material.

Figure~\ref{fig:imagenet_rate_distortion} presents supervised rate-distortion curves of ResNet-50 with various compression approaches, where the x-axis shows the average data size, and the y-axis the supervised performance.
We note that the input tensor shape for ResNet-50 as an image classifier is $3 \times 224 \times 224$.
For image compression, the result shows the considered neural compression models, factorized prior~\cite{balle2018variational} and mean-scale hyperprior~\cite{minnen2018joint}, consistently outperform JPEG and WebP compression in terms of rate-distortion curves in the image classification task.
A popular approach used in split computing studies, the combination of channel reduction and bottleneck quantization (CR+BQ)~\cite{matsubara2020neural}, seems slightly better than JPEG compression but not as accurate as those with the neural image compression models.

Among all the configurations in the figure, our model trained by the two-stage method performs the best. 
We also trained our model without teacher model, which in essence corresponds to~\cite{singh2020end}. The resulting RD curve is significantly worse, which we attribute to two possible effects: first, it is widely acknowledged that knowledge distillation generally finds solutions that generalize better. Second, having a stochastic bottleneck at an earlier layer may make it difficult for the end-to-end training approach to optimize.

\begin{figure}[t]
    \centering
    \includegraphics[width=\linewidth]{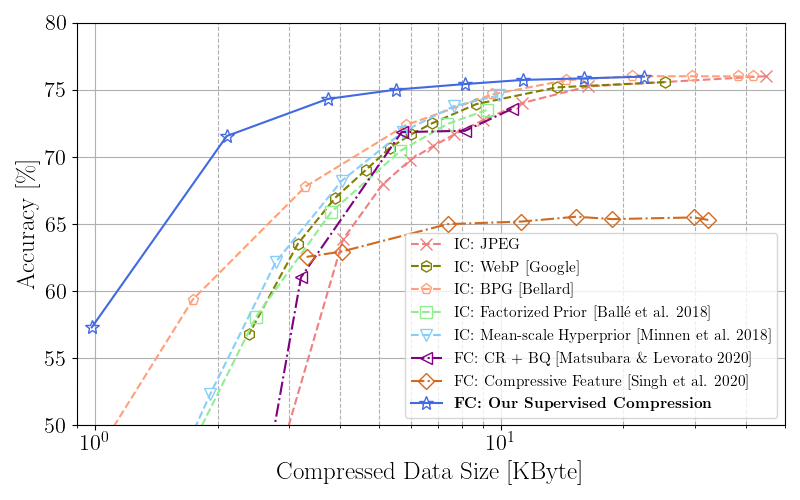}
    \vspace{-2.0em}
    \caption{Rate-distortion (accuracy) curves of ResNet-50 as base model for ImageNet (ILSVRC 2012).}
    \label{fig:imagenet_rate_distortion}
    \vspace{0.5em}
    \includegraphics[width=\linewidth]{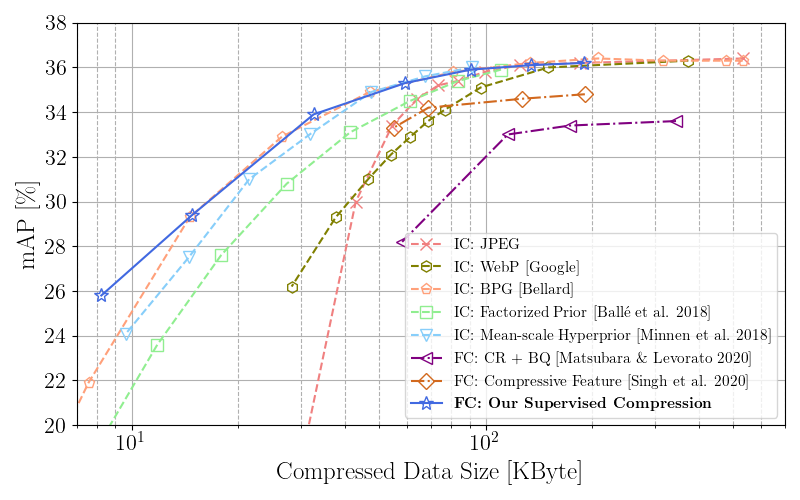}
    \vspace{-2.0em}
    \caption{Rate-distortion (BBox mAP) curves of RetinaNet with ResNet-50 and FPN as base backbone for COCO 2017.}
    \label{fig:coco_bbox_rate_distortion}
    \vspace{0.5em}
    \includegraphics[width=\linewidth]{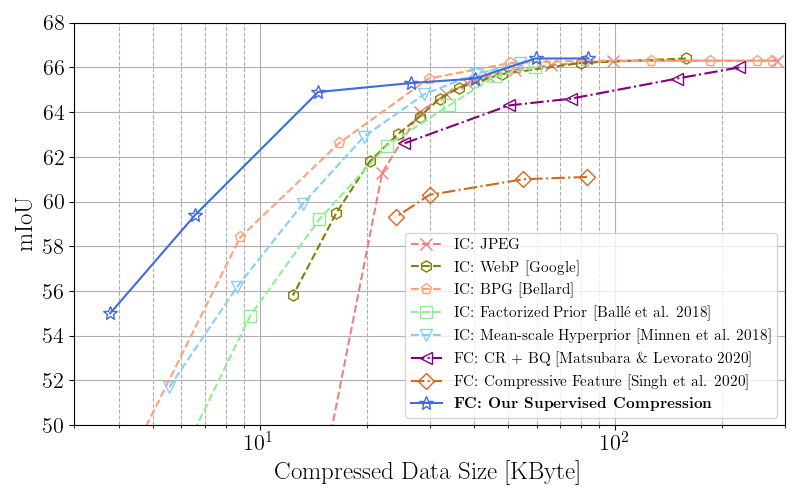}
    \vspace{-2.0em}
    \caption{Rate-distortion (Seg mIoU) curves of DeepLabv3 with ResNet-50 as base backbone for COCO 2017.}
    \label{fig:coco_seg_rate_distortion}
\end{figure}

\subsection{Object Detection and Semantic Segmentation}
\label{subsec:additional_tasks}
As suggested by He \emph{et al.}~\cite{he2019rethinking}, image classifiers pre-trained on the ImageNet dataset~\cite{russakovsky2015imagenet} speed up the convergence of training on downstream tasks.
Reusing the proposed model pre-trained on the ImageNet dataset, we further discuss the rate-distortion performance on two downstream tasks: object detection and semantic segmentation.
Specifically, we train RetinaNet~\cite{tsung2017focal} and DeepLabv3~\cite{chen2017rethinking}, using our models pre-trained on the ImageNet dataset in the previous section as their backbone.
RetinaNet is a one-stage object detection model that enables faster inference than two-stage detectors such as Mask R-CNN~\cite{he2017mask}.
DeepLabv3 is a semantic segmentation model that leverages Atrous Spatial Pyramid Pooling (ASPP)~\cite{chen2017deeplab}.

For the downstream tasks, we use the COCO 2017 dataset~\cite{lin2014microsoft} to fine-tune the models.
The training and validation splits in the COCO 2017 dataset have 118,287 and 5,000 annotated images, respectively.
As detection performance, we refer to mean average precision (mAP) for bounding box (BBox) outputs with different Intersection-over-Unions (IoU) thresholds from 0.5 and 0.95 on the validation split.
For semantic segmentation, we measure the performance by pixel IoU averaged over 21 classes present in the PASCAL VOC 2012 dataset.
It is worth noting that following the PyTorch~\cite{paszke2019pytorch} implementations, the input image scales for RetinaNet~\cite{tsung2017focal} are defined by the shorter image side and set to 800 in this study which is much larger than the input image in the previous image classification task.
As for DeepLabv3~\cite{chen2017rethinking}, we use the resized input images such that their shorter size is 520.
The training setup and hyperparameters used to fine-tune the models are described in the supplementary material.

Similar to the previous experiment for the image classification task, Figures~\ref{fig:coco_bbox_rate_distortion} and~\ref{fig:coco_seg_rate_distortion} show that the combinations of neural compression models and the pre-trained RetinaNet and DeepLabv3, which are still strong baselines in object detection and semantic segmentation tasks. Our model demonstrates better rate-distortion curves in both tasks.
In the object detection task, our model's improvements over RetinaNet with BPG and mean-scale hyperprior are smaller than those in the image classification and semantic segmentation tasks.
However, our model's encoder to be executed on a mobile device is approximately 40 times smaller than the encoder of the mean-scale hyperprior.
Our model also can achieve a much shorter latency to complete the input-to-prediction pipeline (see Fig.~\ref{fig:input_comp_vs_feat_comp}) than the baselines we considered for resource-constrained edge computing systems.
We further discuss these aspects in Sections~\ref{subsec:model_size} and~\ref{subsec:latency_eval}.

\subsection{Bitrate Allocation of Latent Representations}
\begin{figure}[t]
    \centering
    \begin{subfigure}[c]{0.32\linewidth}
    \centering
    \includegraphics[width=1.0\linewidth]{figs/visualizations/000000010764.jpg}
    \label{fig:vis1_input}
    \end{subfigure}
    \begin{subfigure}[c]{0.33\linewidth}
    \centering
    \includegraphics[width=1.0\linewidth]{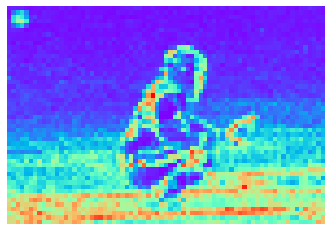}
    \label{fig:vis1_factorized_prior}
    \end{subfigure}
    \begin{subfigure}[c]{0.33\linewidth}
    \centering
    \includegraphics[width=1.0\linewidth]{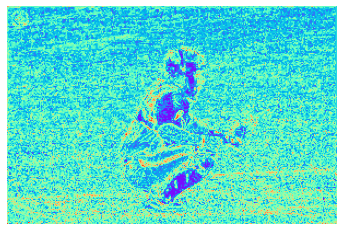}
    \label{fig:vis1_ours}
    \end{subfigure}
    \\\vspace{-1em}
    \centering
    \begin{subfigure}[c]{0.3175\linewidth}
    \centering
    \includegraphics[width=1.0\linewidth]{figs/visualizations/000000447917.jpg}
    \label{fig:vis2_input}
    \end{subfigure}
    \begin{subfigure}[c]{0.33\linewidth}
    \centering
    \includegraphics[width=1.0\linewidth]{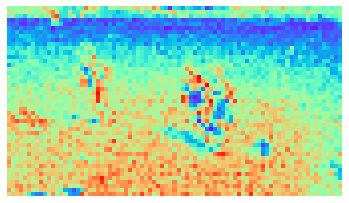}
    \label{fig:vis2_factorized_prior}
    \end{subfigure}
    \begin{subfigure}[c]{0.33\linewidth}
    \centering
    \includegraphics[width=1.0\linewidth]{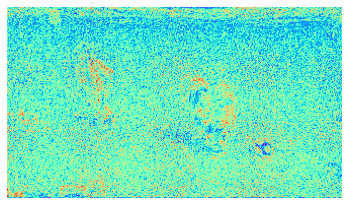}
    \label{fig:vis2_ours}
    \end{subfigure}
    \\\vspace{-1em}
    \centering
    \begin{subfigure}[t]{0.318\linewidth}
    \centering
    \includegraphics[width=1.0\linewidth]{figs/visualizations/000000038118.jpg}
    \caption{Input Images}
    \label{fig:vis3_input}
    \end{subfigure}
    \begin{subfigure}[t]{0.33\linewidth}
    \centering
    \includegraphics[width=1.0\linewidth,trim={0 0.5em 0 0.5em},clip]{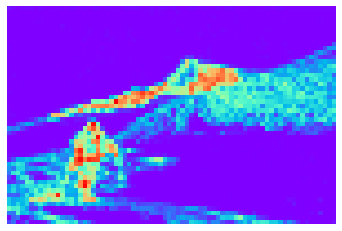}
    \caption{Factorized Prior}
    \label{fig:vis3_factorized_prior}
    \end{subfigure}
    \begin{subfigure}[t]{0.33\linewidth}
    \centering
    \includegraphics[width=1.0\linewidth,trim={0 0.6em 0 0.6em},clip]{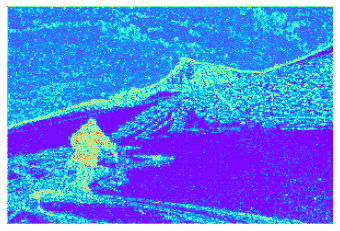}
    \caption{\uourstudent}
    \label{fig:vis3_ours}
    \end{subfigure}
    \\\vspace{-0.5em}
    \caption{Bitrate allocations of latent representations $\z$ in neural image compression and our \ourstudent models. Red and blue areas are allocated higher and lower bitrates, respectively (best viewed in PDF). It appears that the supervised approach (right) allocates more bits to the information relevant to the supervised classification goal.}
    \label{fig:bit_allocations}
\end{figure}

This section discusses the difference between the representations of bottlenecks in neural image compression and our models.
We are interested in which element of the bottlenecks allocates more bits in the latent representation $\z$.
Bottlenecks in neural image compression models will allocate many bits to some \emph{unique} area in an image to preserve all its characteristics in the reconstructed image.
On the other hand, those in our models are trained to mimic the feature representations in their teacher model, thus expected to allocate more bits to areas useful for the target task.

Figure~\ref{fig:bit_allocations} shows visualizations of the normalized bitrate allocations for a few sample images.
The 2nd column of the figure corresponds to the bottleneck in a neural image compression model prioritizing the images' backgrounds such as catcher's zone and glasses.
Interestingly, our bottleneck representation (the 3rd column) seems to eliminate the difference between the two backgrounds and focuses on objects in the images such as persons and soccer ball.
Moreover, the bottleneck eliminates a digital watermark at the top left in the first image, which is most likely not critical for the target task, while the one in the neural compression model noticeably distinguishes the logo from the background.

\subsection{Deployment Cost on Mobile Devices}
\label{subsec:model_size}
In Sections~\ref{subsec:imagenet} and \ref{subsec:additional_tasks}, we discussed the trade-off between transferred data size and model accuracy with the visualization of rate-distortion curves.
While improving the rate-distortion curves is essential for resource-constrained edge computing systems, it is also important to reduce the computational burden allocated to the mobile device that often have more severe constraints on computing and energy resources compared to edge servers.
We investigate then the cost of deploying image classification models on constrained mobile devices.

\begin{table}[t]
    \caption{Number of parameters in compression and classification models loaded on mobile device and edge server. Local (LC), Edge (EC), and Split computing (SC).}
    \vspace{-1em}
    \label{table:model_size}
    \bgroup
    \setlength{\tabcolsep}{0.2em}
    \begin{center}
        \footnotesize
        \begin{tabular}{|c|c|c|c|}
            \hline
            \multicolumn{1}{|c|}{\bf Compression model} & \multicolumn{1}{|c|}{\bf Scenario} & \multicolumn{2}{c|}{\bf Model size (\# params)}\\ \hline
            \hline
            Factorized Prior~\cite{balle2018variational} & EC & Mobile: 1.30M & Edge: 1.30M+$^*$  \\\hline
            Mean-Scale Hyperprior~\cite{minnen2018joint} & EC & Mobile: 5.53M & Edge: 4.49M+$^*$ \\ \hline \hline
            \multicolumn{1}{|c|}{\bf Classification model} & \multicolumn{1}{|c|}{\bf Scenario} & \multicolumn{2}{c|}{\bf Model size (\# params)}\\ \hline
            \hline
            MobileNetV2~\cite{sandler2018mobilenetv2} & LC & \multicolumn{2}{c|}{3.50M} \\ \hline
            MobileNetV3~\cite{howard2019searching} & LC & \multicolumn{2}{c|}{5.48M} \\ \hline
            ResNet-50~\cite{he2016deep} & EC & \multicolumn{2}{c|}{25.6M} \\ \hline \hline
            ResNet-50 w/ BQ~\cite{matsubara2020neural} & SC & Mobile: 0.01M & Edge: 27.3M \\ \hline
            Our \uourstudent & SC & Mobile: 0.14M & Edge: 26.5M \\ \hline
        \end{tabular}
        \\\footnotesize \raggedright * Size of classification model for EC should be additionally considered.
    \end{center}
    \vspace{-1.5em}
    \egroup
\end{table}

Table~\ref{table:model_size} summarizes numbers of parameters used to represent models deployed on the mobile devices and edge servers under different scenarios.
For example, in the edge computing (EC) scenario, the input data compressed by the compressor of an input compression model is sent to the edge server. The decompressor will reconstruct the input data to complete the inference task with a full classification model.
Thus, only the compressor in the input compression model is accounted for in the computation cost on the mobile device.
In Section~\ref{sec:experiments}, the two input compression models, factorized prior~\cite{balle2018variational} and mean-scale hyperprior~\cite{minnen2018joint}, are strong baseline approaches, and mean-scale hyperprior outperforms the factorized prior in terms of rate-distortion curve.
However, its model size is comparable to or more expensive than popular lightweight models such as MobileNetV2~\cite{sandler2018mobilenetv2} and MobileNetV3~\cite{howard2019searching}.
For this reason, this strategy is not advantageous unless the model deployed on the edge server can offer much higher accuracy than the lightweight models on the mobile device.

In contrast, split computing (SC) models, including our \ourstudent model, perform in-network feature compression while extracting features from the target task's input sample.
As shown in Table~\ref{table:model_size}, the encoder of our model is much smaller (about 10 -- 40 times smaller than) compared to those of the input compression models and the lightweight classifiers.
Moreover, the encoder in our student model can be shared with RetinaNet and DeepLabv3 for different tasks.
When a mobile device has multiple tasks such as image classification, object detection, and semantic segmentation, the single encoder is on memory and executed for an input sample.
We note that ResNet-50 models with channel reduction and bottleneck quantization~\cite{matsubara2020neural} and those for compressive feature~\cite{singh2020end} in their studies require a non-shareable encoder for different tasks.
With their approaches, there are three individual encoders on the memory of the more constrained mobile device, which leads to approximately 3 times larger deployment cost.

\subsection{End-to-End Prediction Latency Evaluation}
\label{subsec:latency_eval}
To compare the prediction latency with the different approaches, we deploy the encoders on two different mobile devices: Raspberry Pi 4 (RPI4) and NVIDIA Jetson TX2 (JTX2).
As an edge server (ES), we use a desktop computer with an NVIDIA GeForce RTX 2080 Ti, assuming the use of LoRa~\cite{samie2016iot} for low-power communications (maximum data rate is 37.5 Kbps).
For all the considered approaches, we use the data points with about 74\% accuracy in Fig.~\ref{fig:imagenet_rate_distortion}, and the end-to-end latency is the sum of 1) execution time to encode an input image on RPI4/JTX2, 2) delay to transfer the encoded data from RPI4/JTX2 to ES, and 3) execution time to decode the compressed data and complete inference on ES.

\begin{table}[tb]
    \caption{End-to-end latency to complete input-to-prediction pipeline for resource-constrained edge computing systems illustrated in Fig.~\ref{fig:input_comp_vs_feat_comp}, using RPI4/JTX2, LoRa and ES. The breakdowns are available in the supplementary material.}
    \vspace{-1em}
    \label{table:latency}
    \bgroup
    \setlength{\tabcolsep}{0.3em}
    \begin{center}
        \footnotesize
        \begin{tabular}{|c|r|r|}
            \hline
            \multicolumn{1}{|c|}{\bf Approach} & \multicolumn{1}{|c|}{\bf RPI4 $\longrightarrow$ ES} & \multicolumn{1}{c|}{\bf JTX2 $\longrightarrow$ ES}\\ \hline
            \hline
            JPEG + ResNet-50 & 2.35 sec & 2.34 sec \\ \hline
            WebP + ResNet-50 & 1.83 sec & 1.84 sec \\ \hline
            BPG + ResNet-50 & 2.46 sec & 2.41 sec \\ \hline
            Factorized Prior + ResNet-50 & 2.43 sec & 2.22 sec \\ \hline
            Mean-Scale Hyperprior + ResNet-50 & 2.24 sec & 1.92 sec \\ \hline
            ResNet-50 w/ BQ & 2.27 sec & 2.25 sec \\ \hline \hline
            Our \uourstudent & \textbf{0.972 sec} & \textbf{0.904 sec} \\ \hline
        \end{tabular}
    \end{center}
    \egroup
\end{table}

Table~\ref{table:latency} shows that our approach reduces the end-to-end prediction latency by 47 -- 62\% compared to the baselines.
The encoding time and communication delay are dominant in the end-to-end latency while the execution time on ES is negligible.
For both the experimental configurations (RPI4 $\longrightarrow$ ES and JTX2 $\longrightarrow$ ES), the breakdowns of the end-to-end latency are illustrated in the supplementary material.

\section{Conclusions}
\label{sec:conclusion}
This paper adopts ideas from knowledge distillation and neural image compression to achieve feature compression for supervised tasks.
Our approach leverages a teacher model to introduce a stochastic bottleneck and a learnable prior for entropy coding at its early stage of a student model (namely, \emph{\uourstudent}).
The framework reduces the computational burden on the weak mobile device by offloading most of the computation to a computationally powerful cloud/edge server, and the single encoder in our \ourstudent can serve multiple downstream tasks.
The experimental results show the improved supervised rate-distortion performance for three different vision tasks and the shortened end-to-end prediction latency, compared to various (neural) image compression and feature compression baselines.

To ensure reproducibility of the experimental results and facilitate research to address this important problem, we release the training code and trained models at \repourl.

\subsection*{Acknowledgements}
This material is in part based upon work supported by the Defense Advanced Research Projects Agency (DARPA) under Contract No. HR001120C0021. Any opinions, findings and conclusions or recommendations expressed in this material are those of the author(s) and do not necessarily reflect the views of DARPA. We acknowledge the support by the Google Cloud Platform research credits, and the National Science Foundation under the NSF CAREER award 2047418 and Grants 1928718, 2003237, 2007719, IIS-1724331 and MLWiNS-2003237, as well as Intel and Qualcomm. We also thank Yibo Yang and the anonymous reviewers for their insightful comments. 

{\small
\bibliographystyle{ieee_fullname}
\bibliography{refs}
}

\section*{Supplementary Material}
\appendix
\section{Image Compression Codecs}
As image compression baselines, we use JPEG, WebP~\cite{webp}, and BPG~\cite{bpg}.
For JPEG and WebP, we follow the implementations in Pillow\footnote{\url{https://python-pillow.org/}} and investigate the rate-distortion (RD) tradeoff for the combination of the codec and pretrained downstream models by tuning the quality parameter in range of 10 to 100.
Since BPG is not available in Pillow, our implementation follows~\cite{bpg} and we tune the quality parameter in range of 0 to 50 to observe the RD curve.
We use the x265 encoder with 4:4:4 subsampling mode and 8-bit depth for YCbCr color space, following~\cite{begaint2020compressai}.

\section{Quantization}
\label{sec:quantization}
This section briefly introduces the quantization technique used in both proposed methods and neural baselines with entropy coding.

\subsection{Encoder and Decoder Optimization}
As entropy coding requires discrete symbols, we leverage the method that is firstly proposed in~\cite{balle2017end} to learn a discrete latent variable. During the training stage, the quantization is simulated with a uniform noise to enable gradient-based optimization:
\begin{equation}
    \z=f_\theta(\x) + \mathcal{U}(-\frac{1}{2}, \frac{1}{2}).
\end{equation}
During the inference session, we round the encoder output to the nearest integer for entropy coding and the input of the decoder:
\begin{equation}
    \z=\round{f_\theta(\x)}.
\end{equation}

\subsection{Prior Optimization}
For entropy coding, a prior that can precisely fit the distribution of the latent variable reduces the bitrate. However, the prior distributions such as Gaussian and Logistic distributions are continuous, which is not directly compatible with discrete latent variables. Instead, we use the cumulative of a continuous distribution to approximate the probability mass of a discrete distribution. \cite{balle2017end}:
\begin{equation}
    P(\z)=\int_{\z-\frac{1}{2}}^{\z+\frac{1}{2}} p(t)\text{d}t,
\end{equation}
\noindent where $p$ is the prior distribution we choose, and $P(\z)$ is the corresponding probability mass under the discrete distribution $P$. The integral can easily be computed with the Cumulative Distribution Function (CDF) of the continuous distribution.

\section{Neural Image Compression}
In this section, we describe the experimental setup that we used for the neural image compression baselines.

\subsection{Network Architecture}
\paragraph{Factorized prior model~\cite{balle2018variational}.}
This model consists of 4 convolutional layers for encoding and 4 deconvolutional layers for decoding. Each layer follows (128, 5, 2, 2) configuration in the format (number of channels, kernel size, stride, padding). We also use the simplified version of generalized divisive normalization (GDN) and inversed GDN (IGDN)~\cite{balle2015density} as activation functions for the encoder and decoder, respectively. The prior distribution uses a univariate non-parametric density model, whose cumulative distribution is parameterized by a neural network~\cite{balle2018variational}.

\paragraph{Mean-scale hyperprior model.}
We use exactly the same architecture described in~\cite{minnen2018joint}.

\subsection{Training}
All the models are trained on a high-resolution dataset with around 2,700 images collected from DIV2K dataset~\cite{agustsson2017ntire} and CLIC dataset~\cite{clic2020}. During training, we apply random crop size (256, 256) to the images and set the batch size as 8. We also use Adam~\cite{kingma2015adam} optimizer with $10^{-4}$ learning rate to train the model for 900,000 steps, and then the learning rate is decayed to $10^{-5}$ for another 100,000 steps.

\section{Channel Reduction and Bottleneck\\Quantization}
A combination of channel reduction and bottleneck quantization (CR + BQ) is a popular approach in studies on split computing~\cite{eshratifar2019bottlenet,matsubara2020head,shao2020bottlenet++,matsubara2020neural}, and we refer to the approach as a baseline.

\begin{figure*}[t]
    \centering
    \includegraphics[width=0.85\linewidth]{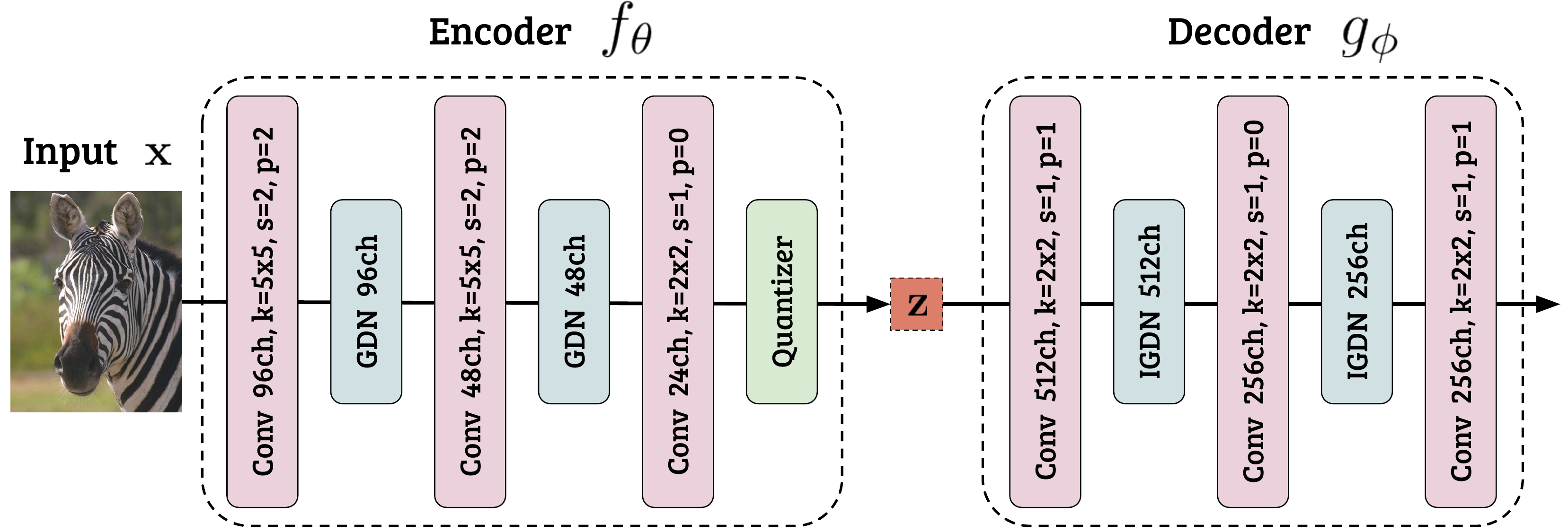}
    \caption{Our encoder and decoder introduced to ResNet-50. k: kernel size, s: stride, p: padding.}
    \label{fig:encoder_decoder}
\end{figure*}

\subsection{Network Architecture}
\paragraph{Image classification.}
We reuse the architectures of encoder and decoder from Matsubara \emph{et al.}~\cite{matsubara2020head} introduced in ResNet~\cite{he2016deep} and validated on the ImageNet (ILSVRC 2012) dataset~\cite{russakovsky2015imagenet}.
Following the study, we explore the rate-distortion (RD) tradeoff by varying the number of channels in a convolution layer (2, 3, 6, 9, and 12 channels) placed at the end of the encoder and apply a quantization technique (32-bit floating point to 8-bit integer)~\cite{jacob2018quantization} to the bottleneck after the training session.

\paragraph{Object detection and semantic segmentation.}
Similarly, we reuse the encoder-decoder architecture used as ResNet-based backbone in Faster R-CNN~\cite{ren2015faster} and Mask R-CNN~\cite{he2017mask} for split computing~\cite{matsubara2020neural}.
The same ResNet-based backbone is used for RetinaNet~\cite{tsung2017focal} and DeepLabv3~\cite{chen2017rethinking}.
Again, we examine the RD tradeoff by controlling the number of channels in a bottleneck layer (1, 2, 3, 6, and 9 channels) and apply the same post-training quantization technique~\cite{jacob2018quantization} to the bottleneck.

\subsection{Training}
Using ResNet-50~\cite{he2016deep} pretrained on the ImageNet dataset as a teacher model, we train the encoder-decoder introduced to a copy of the teacher model, that is treated as a student model for image classification.
We apply the generalized head network distillation (GHND)~\cite{matsubara2020neural} to the introduced encoder-decoder in the student model.
The model is trained on the ImageNet dataset to mimic the intermediate features from the last three residual blocks in the teacher (ResNet-50) by minimizing the sum of squared error losses.
Using the Adam optimizer~\cite{kingma2015adam}, we train the student model on the ImageNet dataset for 20 epochs with the training batch size of 32.
The initial learning rate is set to $10^{-3}$ and reduced by a factor of 10 at the end of the 5th, 10th, and 15th epochs.

Similarly, we use ResNet-50 models in RetinaNet with FPN and DeepLabv3 pretrained on COCO 2017 dataset~\cite{lin2014microsoft} as teachers, and apply the GHND to the students for the same dataset.
The training objective, the initial learning rate, and the number of training epochs are the same as those for the classification task.
We set the training batch size to 2 and 8 for object detection and semantic segmentation tasks, respectively.
The learning rate is reduced by a factor of 10 at the end of the 5th and 15th epochs.

\section{Proposed Student Model}
This section presents the details of student models and training methods we propose in this study.

\subsection{Network Architecture}
As illustrated in Fig.~\ref{fig:encoder_decoder}, our encoder $f_\theta$ is composed of convolution and GDN~\cite{balle2015density} layers followed by a quantizer described in Section~\ref{sec:quantization}.
Similarly, our decoder $g_\phi$ is designed with convolution and inversed GDN (IGDN) layers to have the output tensor shape match that of the first residual block in ResNet-50~\cite{he2016deep}.
For image classification, the entire architecture of our \ourstudent model consists of the encoder and decoder followed by the last three residual blocks, average pooling, and fully-connected layers in ResNet-50.
For object detection and semantic segmentation, we replace ResNet-50 (used as a backbone) in RetinaNet~\cite{tsung2017focal} and DeepLabv3~\cite{chen2017rethinking} with our student model for image classification.

\subsection{Two-stage Training}
Here, we describe the two-stage method we proposed to train the \ourstudent models.
\paragraph{Image classification.}
Using the ImageNet dataset, we put our focus on the introduced encoder and decoder at the first stage of training and then freeze the encoder to fine-tune all the subsequent layers at the second stage for the target task.
At the 1st stage, we train the student model for 10 epochs to mimic the behavior of the first residual block in the teacher model (pretrained ResNet-50) in a similar way to~\cite{matsubara2020neural} but with the rate term to learn a prior for entropy coding.
We use Adam optimizer with batch size of 64 and an initial learning rate of $10^{-3}$.
The learning rate is decreased by a factor of 10 after the end of the 5th and 8th epochs.

Once we finish the 1st stage, we fix the parameters of the encoder that has learnt compressed features at the 1st stage and fine-tune all the other modules, including the decoder for the target task.
By freezing the encoder's parameters, we can reuse the encoder for different tasks. The rest of the layers can be optimized to adopt the compressible features for the target task. Note that once the encoder is frozen, we also no longer optimize both the prior and encoder, which means we can directly use \emph{rounding} to quantize the latent variable.
With the encoder frozen, we apply a standard knowledge distillation technique~\cite{hinton2014distilling} to achieve better model accuracy, and the concrete training objective is formulated as follows:
\begin{equation}
    \mathcal{L} = \alpha \cdot \mathcal{L}_\text{cls}(\mathbf{\hat{y}}, \mathbf{y}) + (1 - \alpha) \cdot \tau^2 \cdot \mathcal{L}_\text{KL} \left(\mathbf{o}^\text{S}, \mathbf{o}^\text{T}\right),
    \label{eq:proposed_2nd}
\end{equation}
\noindent where $\mathcal{L}_\text{cls}$ is a standard cross entropy.
$\hat{\y}$ indicates the model's estimated class probabilities, and $\y$ is the annotated object category.
$\alpha$ and $\tau$ are both hyperparameters, and $\mathcal{L}_\text{KL}$ is the Kullback-Leibler divergence.
$\mathbf{o}^\text{S}$ and $\mathbf{o}^\text{T}$ represent the \emph{softened} output distributions from student and teacher models, respectively.
Specifically, $\mathbf{o}^\text{S} = [o_1^\text{S}, o_2^\text{S}, \ldots, o_{|\mathcal{C}|}^\text{S}]$ where $\mathcal{C}$ is a set of object categories considered in target task.
$o_i^\text{S}$ indicates the student model's softened output value (scalar) for the $i$-th object category:
\begin{equation}
    o_{i}^\text{S} = \frac{\exp \left( \frac{v_i}{\tau} \right)}{\sum_{k \in \mathcal{C}} \exp\left( \frac{v_k}{\tau} \right)},
\end{equation}
\noindent where $\tau$ is a hyperparameter defined in Eq.~\ref{eq:proposed_2nd} and called \emph{temperature}.
$v_i$ denotes a logit value for the $i$-th object category. 
The same rules are applied to $\mathbf{o}^\text{T}$ for teacher model.

For the 2nd stage, we use the stochastic gradient descent (SGD) optimizer with an initial learning rate of $10^{-3}$, momentum of 0.9, and weight decay of $5 \times 10^{-4}$.
We reduce the learning rate by a factor of 10 after the end of the 5th epoch, and the training batch size is set to 128.
The balancing weight $\alpha$ and temperature $\tau$ for knowledge distillation are set to 0.5 and 1, respectively.

\paragraph{Object detection.}
We reuse the \ourstudent model trained on the ImageNet dataset in place of ResNet-50 in RetinaNet~\cite{tsung2017focal} and DeepLabv3~\cite{chen2017rethinking} (teacher models).
Note that we freeze the parameters of the encoder trained on the ImageNet dataset to make the encoder sharable for multiple tasks.
Reusing the encoder trained on the ImageNet dataset is a reasonable approach as 1) the ImageNet dataset contains a larger number of training samples (approximately 10 times more) than those in the COCO 2017 dataset~\cite{lin2014microsoft}; 2) models using an image classifier as their backbone frequently reuse model weights trained on the ImageNet dataset~\cite{ren2015faster,tsung2017focal}.

To adapt the encoder for object detection, we train the decoder for 3 epochs at the 1st stage in the same way we train those for image classification (but with the encoder frozen).
The optimizer is Adam~\cite{kingma2015adam}, and the training batch size is 6.
The initial learning rate is set to $10^{-3}$ and reduced to $10^{-4}$ after the first 2 epochs.
At the 2nd stage, we fine-tune the whole model except its encoder for 2 epochs by the SGD optimizer with learning rates of $10^{-3}$ and $10^{-4}$ for the 1st and 2nd epochs, respectively.
We set the training batch size to 6 and follow the training objective in~\cite{tsung2017focal}, which is a combination of L1 loss for bounding box regression and Focal loss for object classification.

\paragraph{Semantic segmentation.}
For semantic segmentation, we train DeepLabv3 in a similar way.
At the 1st stage, we freeze the encoder and train the decoder for 5 epochs, using Adam optimizer with batch size of 8.
The initial learning rate is $10^{-3}$ and decreased to $10^{-4}$ after the first 3 epochs.
At the 2nd stage, we train the entire model except for its encoder for 5 epochs.
We minimize a standard cross entropy loss, using the SGD optimizer.
The initial learning rates for the body and the sub-branch (auxiliary module)\footnote{\url{https://github.com/pytorch/vision/tree/master/references/segmentation}} are $2.5 \times 10^{-3}$ and $2.5 \times 10^{-2}$, respectively.
Following~\cite{chen2017rethinking}, we reduce the learning rate after each iteration as follows:
\begin{equation}
    lr = lr_0 \times \left(1 - \frac{N_\text{iter}}{N_\text{max\_iter}}\right)^{0.9},
    \label{eq:seg_lr}
\end{equation}
\noindent where $lr_0$ is the initial learning rate.
$N_\text{iter}$ and $N_\text{max\_iter}$ indicate the accumulated number of iterations and the total number of iterations, respectively.

\subsection{End-to-end Training}
In this work, the end-to-end training approach for feature compression~\cite{singh2020end} is treated as a baseline and applied to our \ourstudent model without teacher models.

\begin{figure*}[t]
    \centering
    \includegraphics[width=0.95\linewidth]{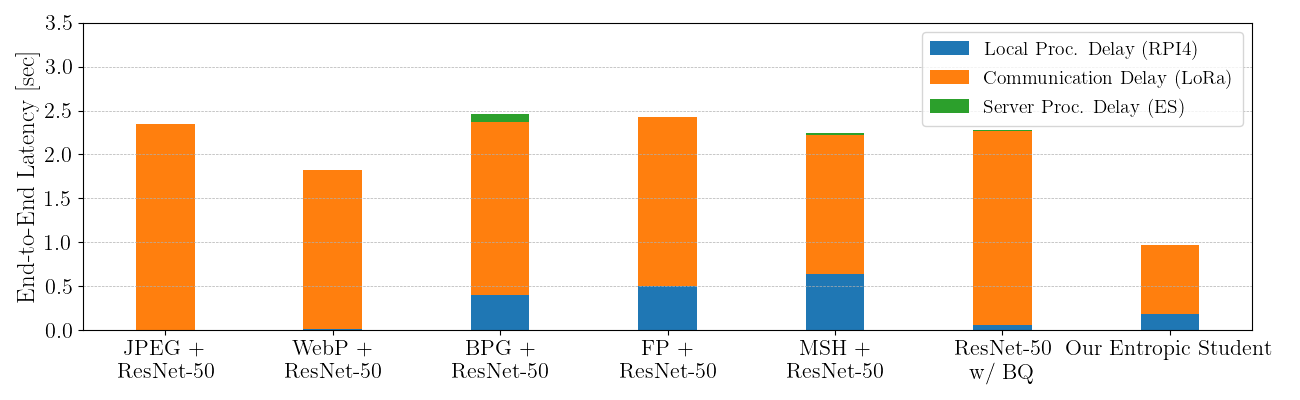}
    \vspace{-1em}
    \caption{Component-wise delays to complete input-to-prediction pipeline, using \uline{RPI4} as mobile device.}
    \label{fig:rpi4_es_latency}
    \includegraphics[width=0.95\linewidth]{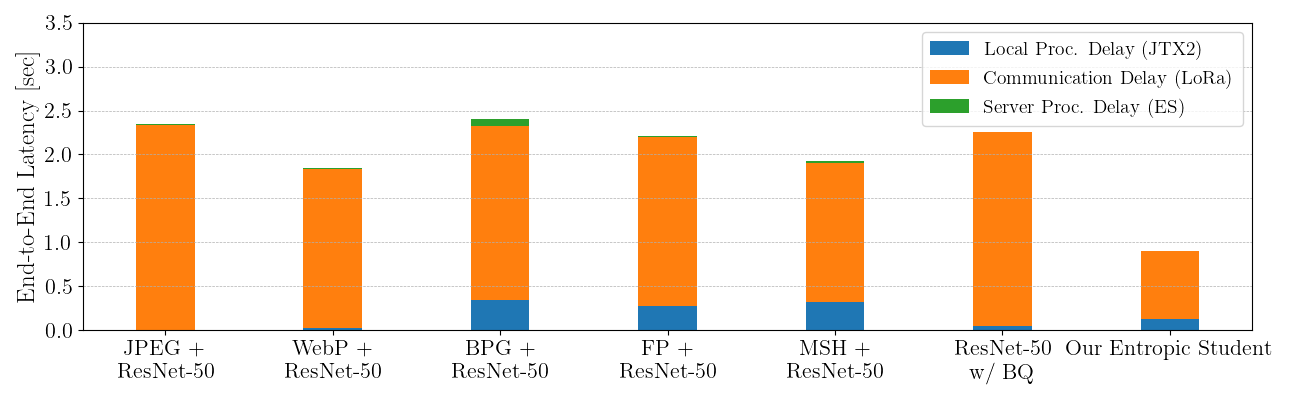}
    \vspace{-1em}
    \caption{Component-wise delays to complete input-to-prediction pipeline, using \uline{JTX2} as mobile device.}
    \label{fig:jtx2_es_latency}
\end{figure*}

\paragraph{Image classification.}
Following the end-to-end training approach~\cite{singh2020end}, we train our \ourstudent model from scratch.
Specifically, we use Adam~\cite{kingma2015adam} optimizer and cosine decay learning rate schedule~\cite{loshchilov17sgdr} with an initial learning rate of $10^{-3}$ and weight decay of $10^{-4}$.
Based on their training objectives (Eq.~\ref{eq:end2end}), we train the model for 60 epochs with batch size of 256.\footnote{For the ImageNet dataset, Singh \emph{et al.} train their models for 300k steps with batch size of 256 for 1.28M training samples, which is equivalent to 60 epochs ($= \frac{300k \times 256}{1.28M}$).}
Note that Singh \emph{et al.}~\cite{singh2020end} evaluate the accuracy of their models on a $299 \times 299$ center crop.
Since the pretrained ResNet-50 expects the crop size of $224 \times 224$,\footnote{\url{https://pytorch.org/vision/stable/models.html\#classification}} we use the crop size for all the considered classifiers to highlight the effectiveness of our approach.

\begin{equation}
    \mathcal{L} = \underbrace{\mathcal{L}_\text{cls}(\mathbf{\hat{y}}, \mathbf{y})}_\text{distortion} \\ - \beta \underbrace{\log p_\phi(f_\theta(\x) + \epsilon)}_{\text{rate}}, \quad \epsilon \sim \text{Unif}(\textstyle{-\frac{1}{2},\frac{1}{2}})
    \label{eq:end2end}
\end{equation}

\paragraph{Object detection.}
Reusing the model trained on the ImageNet dataset with the end-to-end training method, we fine-tune RetinaNet~\cite{tsung2017focal}.
Since we empirically find that a standard transfer learning approach\footnote{\url{https://github.com/pytorch/vision/tree/master/references/detection}} to RetinaNet with the model trained by the baseline method did not converge, we apply the 2nd stage of our fine-tuning method described above to the RetinaNet model.
The hyperparameters are the same as above, but the number of epochs for the 2nd stage training is 5.

\paragraph{Semantic segmentation.}
We fine-tune DeepLabv3~\cite{chen2017rethinking} with the same model trained on the ImageNet dataset.
Using the SGD optimizer with an initial learning rate of 0.01, momentum of 0.9, and weight decay of 0.001, we minimize a standard cross entropy loss.
The learning rate is adjusted by Eq.~\ref{eq:seg_lr}, and we train the model for 30 epochs with batch size of 16.

\section{End-to-End Prediction Latency}

In this section, we provide the detail of the end-to-end prediction latency evaluation shown in this work.
Figures~\ref{fig:rpi4_es_latency} and \ref{fig:jtx2_es_latency} show the breakdown of the end-to-end latency per image for Raspberry Pi 4 (RPI4) and NVIDIA Jetson TX2 (JTX2) as mobile devices, respectively.
For each of the configurations we considered, we present 1) local processing delay (encoding delay on mobile device), 2) communication delay to transfer the encoded (compressed) data to edge server by LoRa~\cite{samie2016iot}, and 3) server processing delay to decode the data transferred from mobile device and complete the inference pipeline on edge server (ES).
Following~\cite{matsubara2019distilled,matsubara2020head,matsubara2020neural}, we compute the communication delay by dividing transferred data size by the available data rate, 37.5 Kbps (LoRa~\cite{samie2016iot}) in this paper.
For all the considered approaches, we use the data points with about 74\% accuracy in our experiments with the ImageNet dataset.

From the figures, we can confirm that the communication delay is dominant in the end-to-end latency for all the approaches we considered, and the third component (server processing delay) is also negligible as the edge server has more computing power that the mobile devices have.
Overall, our \ourstudent model successfully saves the end-to-end prediction latency by compressing the data to be transferred to edge server with a small portion of computing cost on mobile device.

\end{document}